\documentclass[runningheads]{llncs}
\usepackage{graphicx}

\usepackage{tikz}
\usepackage{comment}
\usepackage{amsmath,amssymb} 
\usepackage{hyperref}
\usepackage{color}

\usepackage[accsupp]{axessibility}  

\usepackage{multirow}
\usepackage{makecell}
\usepackage{algorithm}
\usepackage{algorithmicx}
\usepackage{algpseudocode}
\begin{document}
\pagestyle{headings}
\mainmatter
\def\ECCVSubNumber{1814}  

\title{Learning Spatiotemporal Frequency-Transformer for Compressed Video Super-Resolution} 


\titlerunning{Spatiotemporal Frequency-Transformer}
%
\author{Zhongwei Qiu\inst{1,2*} \and
Huan Yang\inst{3} \and
Jianlong Fu\inst{3} \and Dongmei Fu\inst{1,2}}
\authorrunning{Z. Qiu, H. Yang, J. Fu, and D. Fu}
%
\institute{University of Science and Technology Beijing \and
Shunde Graduate School of University of Science and Technology Beijing \and
Microsoft Research\\ \email{qiuzhongwei@xs.ustb.edu.cn}, \email{huayan@microsoft.com}, \email{jianf@microsoft.com}, \email{fdm\_ustb@ustb.edu.cn}}

\maketitle

\begin{abstract}
   \footnotetext[1]{This work was done when Z. Qiu was an intern at Microsoft Research.}

Compressed video super-resolution (VSR) aims to restore high-resolution frames from compressed low-resolution counterparts. Most recent VSR approaches often enhance an input frame by ``borrowing’’ relevant textures from neighboring video frames. Although some progress has been made, there are grand challenges to effectively extract and transfer high-quality textures from compressed videos where most frames are usually highly degraded. In this paper, we propose a novel Frequency-Transformer for compressed video super-resolution (FTVSR) that conducts self-attention over a joint space-time-frequency domain. First, we divide a video frame into patches, and transform each patch into DCT spectral maps in which each channel represents a frequency band. Such a design enables a fine-grained level self-attention on each frequency band, so that real visual texture can be distinguished from artifacts, and further utilized for video frame restoration. Second, we study different self-attention schemes, and discover that a ``divided attention'' which conducts a joint space-frequency attention before applying temporal attention on each frequency band, leads to the best video enhancement quality. Experimental results on two widely-used video super-resolution benchmarks show that FTVSR outperforms state-of-the-art approaches on both uncompressed and compressed videos with clear visual margins.
Code are available at https://github.com/researchmm/FTVSR.

\end{abstract}

\section{Introduction}

Video super-resolution (VSR) aims to restore a sequence of high-resolution (HR) frames from its low-resolution (LR) counterparts. It is a fundamental computer vision task, and can benefit a broad range of downstream applications, such as video surveillance~\cite{zhang2010super} and high-definition television~\cite{goto2014super}. State-of-the-art VSR approaches mainly focus on leveraging temporal information by sliding windows~\cite{kim20183dsrnet,li2019fast,tian2020tdan,wang2019edvr} or recurrent structures~\cite{chan2021basicvsr,sajjadi2018frame,yi2021omniscient}, and have achieved great success in limited scenarios that usually take uncompressed video frames as inputs.

\begin{figure}[!t]
\centering
\includegraphics[width=\columnwidth]{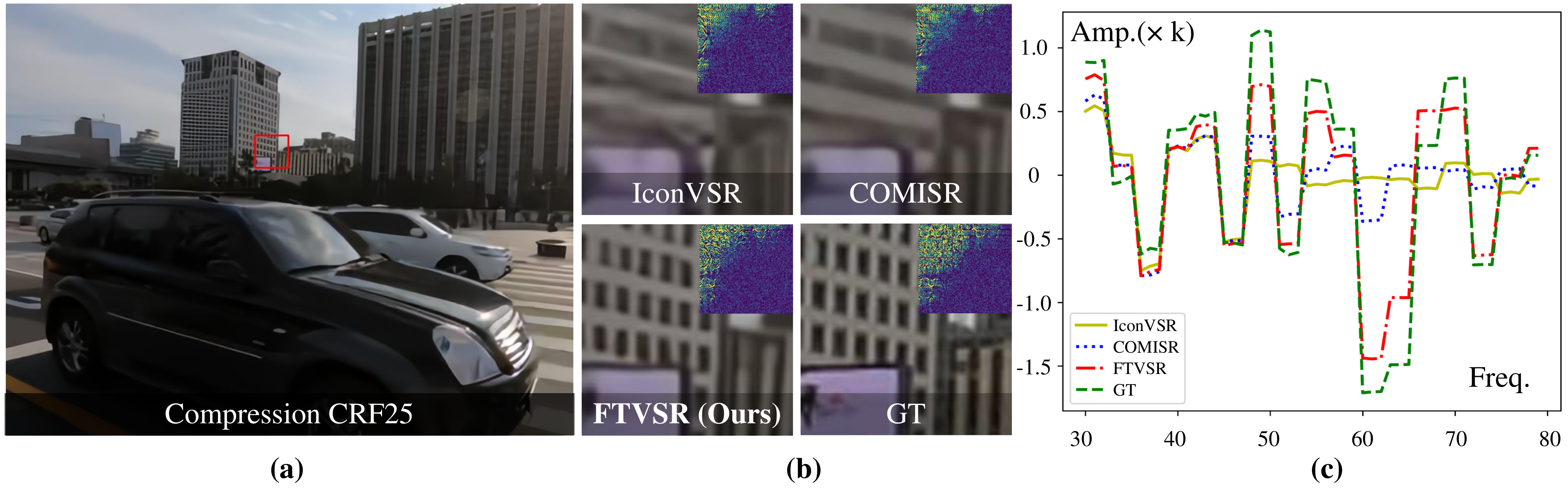}
\caption{Comparison of our FTVSR and state-of-the-art VSR methods (IconVSR~\cite{chan2021basicvsr}, COMISR~\cite{li2021comisr}) on compressed videos with a compression rate of CRF25. (a) The results of $\times 4$ VSR by FTVSR. (b) Comparison of the zoom-in patches and their DCT-based spectral maps (shown in the top-right corner). FTVSR recovers more high-frequency information than IconVSR and COMISR. (c) Comparison of the Amplitude-Frequency curves on clipped frequency bands of 30 to 80. The proposed FTVSR is superior than other methods to approximate the curve of ground-truth (Best viewed in color)}
\label{fig:teaser}
\end{figure}

However, most videos on the internet or in user devices are stored and transmitted in a compressed format. For example, the most widely-used video codec H.264 takes a constant rate factor (CRF) varied from $0$ to $51$ as its parameter to control the compression rate. 
As shown in Figure~\ref{fig:teaser}, directly applying the state-of-the-art IconVSR approach~\cite{chan2021basicvsr} to such a compressed scenario failed to generate visually pleasant results. Because the model trained on uncompressed videos often treats the unseen compression artifacts as common textures and magnifies these artifacts during restoration processes. 

Recent progress has been made by taking into account the compression artifacts in VSR model design. To strength the awareness of compression, a pioneer work, COMISR~\cite{li2021comisr}, is proposed to predict detail-aware flow to align high-resolution features and enhance HR images by Laplacian enhancement module. However, there are still large gaps between the generated frame and the ground-truth, as shown in Figure~\ref{fig:teaser}.  

To solve the above issues, we propose a novel \textbf{F}requency \textbf{T}ransformer for compressed \textbf{V}ideo \textbf{S}uper-\textbf{R}esolution (\textbf{FTVSR}). The key insight is to transform a compressed video frame into a bunch of frequency-based patch representations by Discrete Cosine Transform (DCT), and design frequency-based attention to enable deep feature fusions across multiple frequency bands. Such a design has the following two key merits: 1) the DCT-based representation treats each frequency band ``fairly'', so that high-frequency visual details can be well-preserved; 2) the frequency attention enables low-frequency information (e.g., object structure) to guide the generation of high-frequency textures, so that the effect of compression artifacts can be greatly reduced. Besides, to further utilize the spatial and temporal dependencies in videos, we extensively explore different frequency attention mechanisms that combine with space and time attention in a proper manner. Extensive experiments on two widely-used VSR benchmarks demonstrate that the proposed FTVSR significantly outperforms previous methods and achieves new SOTA results. For example, for the setting of $CRF=25$ in the REDS dataset, the gains of the proposed FTVSR are nearly $1.6$ dB and $2.1$ dB, compared with the competitive COMISR~\cite{li2021comisr} and IconVSR~\cite{chan2021basicvsr}, respectively.

\section{Related Work}
\subsection{Video Super-Resolution}

\subsubsection{Uncompressed Video Super-Resolution}
Modern video super-resolution approaches~\cite{chan2021basicvsr,yi2021omniscient,kim20183dsrnet,tian2020tdan,wang2019edvr,yang2020learning,sajjadi2018frame,liu2022learning} focus on improving the quality of HR sequences by extracting more information from temporal features, which can be categorized into sliding-window  and recurrent structure. 
The approaches~\cite{kim20183dsrnet,tian2020tdan,wang2019edvr} based on sliding-window structure recover HR frames from adjacent LR frames within a sliding-window. They mainly use 3D convolution~\cite{kim20183dsrnet}, optical flow~\cite{kim2018spatio,tao2017detail} or deformable convolution~\cite{tian2020tdan,wang2019edvr} to align the temporal features.
However, these methods can't utilize the temporal features from long-distance frames. 
Other approaches~\cite{tao2017detail,haris2019recurrent,isobe2020video,yi2021omniscient,chan2021basicvsr} based recurrent structure usually use a hidden state to transmit temporal information from long-distance frames.
BasicVSR and IconVSR~\cite{chan2021basicvsr} achieve significant improvements with bidirectional recurrent structure, which fuses the forward and backward propagation features. 
Recently, transformer-based approaches~\cite{zeng2021improving,li2020mucan,cao2021video} make great success by using different attention~\cite{fu2017look,zheng2017learning} to capture temporal features. Limited by computational costs, they just can aggregate information from a few adjacent frames. Despite the remarkable progress by these approaches, they are focus on uncompressed videos and usually fail to recover the HR frames from compressed LR frames.

\subsubsection{Compressed Video Super-Resolution}
Compared with uncompressed VSR, compressed VSR is more difficult due to the lost information and the extra high-frequency artifacts caused by compression. There are three potential solutions to handle the compressed problem: video denoising, training on compressed videos and specific model design for compression. COMISR~\cite{li2021comisr} firstly applies different video denoising~\cite{lu2018deep,lu2019deep,xu2019non} on compressed videos to remove the artifacts and uses state-of-the-art VSR methods~\cite{wang2019edvr,li2020mucan,chan2021basicvsr} on the denoised LR videos. Experimental results have shown that this pre-process is not working since the degradation kernel used for training VSR is different from the denoising model. COMISR~\cite{li2021comisr} further designs detail-aware module to align high-resolution features and Laplacian module to enhance HR frames with recurrent structure. However, these designs can not distinguish the high-frequency textures from the artifacts since these signals are coupled.

\subsection{Frequency Learning}
Lot of studies explore to learn in frequency domain, including high-level semantic tasks~\cite{ehrlich2019deep,xu2020learning,qin2021fcanet} and low-level restoration tasks~\cite{wang2016d3,ehrlich2020quantization,li2021learning,fritsche2019frequency}. High-level semantic tasks usually reduce the computational cost by transforming images into frequency domain. Particularly, FcaNet~\cite{qin2021fcanet} propose frequency channel attention to improve the performance of ResNet on classification task.  Many low-level studies explore to restore content details from frequency decomposition perspective. Parts of them~\cite{fritsche2019frequency,li2021learning} study decomposing features into different frequency bands by multi-branch CNNs. Typically, OR-Net~\cite{li2021learning} uses multi-branch CNNs to separate different frequency components and enhances these features with frequency enhancement unit. Another parts~\cite{wang2016d3,ehrlich2020quantization} of them transform images into frequency domain. For example, D$^3$~\cite{wang2016d3} designs a dual-domain restoration network to remove artifacts of JPEG compressed images. Moreover, Ehrlich \textit{et al.}~\cite{ehrlich2020quantization} designs a Y-channel correction network and a color channel correction network in frequency domain to correct the JPEG artifact. Existing VSR methods are developed in pixel domain, but the video compression problem is generated in frequency domain. Inspired by this, we introduce a frequency-transformer to tackle the compression problems in VSR.

\section{Approach}

\subsection{Problem formulation} 
VSR aims to restore the HR videos from its LR counterparts without taking into account video compression. Our focus, compressed VSR, aims to recover the HR frames from its compressed LR frames, which is more difficult. 
Let $I_{LR} = \{I^t_{LR}|t\in[1,T]\}$ be a compressed LR sequence of height $H$, width $W$, and frame length $T$. 
The restored super-resolution frames are denoted as $I_{SR} = \{I^t_{SR}|t\in[1,T]\}$ of height $\alpha H$, width $\alpha W$, in which
$\alpha$ represents the upsampling scale factor. The corresponding HR frames are denoted as $I_{HR} = \{I^t_{HR}|t\in[1,T]\}$.

\subsection{Frequency-based Tokenization}
\label{dct_tokens}
To solve the problem of compressed video super-resolution, we propose to adopt a frequency-based patch representation. Following the previous works~\cite{gueguen2018faster,ehrlich2019deep,ehrlich2020quantization} in computer vision, we adopt the widely-used method, DCT, as our operation to transfer an image into frequency domain.
 
\begin{figure}
\centering
\includegraphics[width=\columnwidth]{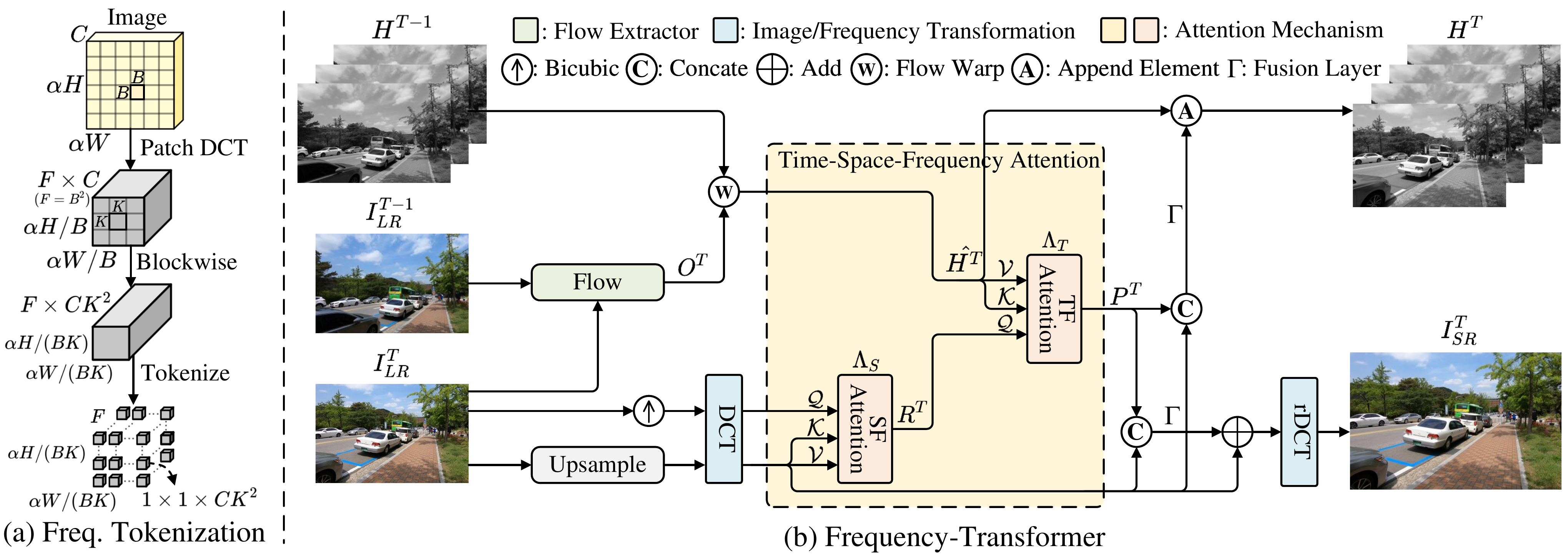}
\caption{(a) An RGB image is extracted as frequency tokens of size $C\times K \times K$ by DCT-based frequency tokenization. (b) A Frequency-Transformer with divide time-space-frequency (TSF) attention $\Lambda_{ST}$, which achieves best performance in our experiments. Given compressed LR sequence, Frequency-Transformer performs TSF attention on the frequency tokens of video frames and output SR frames with a hidden state $H$ maintained by a recurrent structure. TSF consists of $\Lambda_S$ attention and $\Lambda_T$ attention. The $Q$, $K$, $V$ of $\Lambda_S$ are tokens from videos frame sampled $I^T_{LR}$ by bicubic and upsample network, respectively. $R^T$ is the output of $\Lambda_S$, further sums with hidden states $\Hat{H}^T$ warped from past hidden states by flow $O^T$, as the $K$ and $V$ of $\Lambda_T$ attention. $P^T$ is the output of $\Lambda_T$, which further used to update hidden state and recover SR frame $I^T_{SR}$}
\label{fig:FTVSR}
\end{figure}

\subsubsection{DCT} Discrete Cosine Transform projects an image into a set of cosine components for different 2D frequencies. Given an image patch $P$ of height $B$ and width $B$, a $B \times B$ DCT block $D$ is generated as:
\begin{equation}
\label{eq_dct}
    D(u, v) = c(u)c(v)\sum^{B-1}_{x=0}\sum^{B-1}_{y=0}P(x,y)cos[\frac{(2x+1)u\pi}{2B}]cos[\frac{(2y+1)v\pi}{2B}], \\
\end{equation}
where $x$ and $y$ are the 2D indexes of pixels. $u\in [0,B-1]$ and $v\in [0,B-1]$ are the 2D indexes of frequencies. $c(\cdot)$ represents normalizing scale factor to enforce orthonormality and $c(u) = \sqrt{\frac{1}{B}}$ if $u=0$, else $c(u) = \sqrt{\frac{2}{B}}$. The DCT and its inversion are denoted as $\text{DCT}(\cdot)$ and $\text{rDCT}(\cdot)$, respectively.

\subsubsection{DCT-based Frequency Tokenization}
Given a LR sequence, we firstly upsample the $I_{LR}$ by a upsampling network $\varphi(\cdot)$. 
For each frame, we transform each channel of RGB image into frequency domain by applying DCT on the patches of shape $B\times B$ as Equation \ref{eq_dct}, which can be formulated as: 
\begin{equation}
    D_{LR}(u,v) = \text{DCT}(\varphi (I_{LR})),
\end{equation}
where $D_{LR}(u,v)$ of shape $T\times F\times C\times \frac{\alpha H}{B}\times \frac{\alpha W}{B}$ represents the transformed 2D spectral map from LR image. $T$, $F$, $C$, $\frac{\alpha H}{B}$ and $\frac{\alpha W}{B}$ represent sequence length, frequency dimensions, image channels, height and width, respectively. The frequency number is $F = B^2$. 

For a spectral frame $D_{LR}(u,v)$, we split the frequency dimension to form $F$ visual tokens. The frequency tokens set $\mathcal{T}$ can be represented as: 
\begin{equation}
    \mathcal{T} = \{\tau_f, f\in [1, F]\},
\end{equation}
where $\tau_f$ represents the frequency token in $f^{th}$ frequency, which has a feature size of $C\times \frac{\alpha H}{B}\times \frac{\alpha W}{B}$. 
This frequency tokenization mechanism brings the information exchange between different frequency bands and forces neural network treating low-frequency signals and high-frequency signals ``fairly", which is beneficial to preserve high-frequency visual details. Combined with frequency attention mechanism in Section \ref{frequecny_transformer}, the high-frequency textures can be restored well by the guidance of low-frequency information (e.g., object structure).

In order to capture the frequency relationship between different spatial blocks, the spectral maps are split into a set of blocks with a kernel size of $K\times K$. To further extract temporal information, we extend the same tokenization to all video frames.
Therefore, we generate more fine-grained frequency tokens in both space and time dimensions, which can be represented as:
\begin{equation}
    \mathcal{T} = \{\tau_{(t,i,f)}, t\in [1,T], i\in [1,N], f\in [1, F]\},
\end{equation}
where each token $\tau_{(t,i,f)}$ has a shape of $C\times K\times K$. $N$ represents the generated block number in each frame. Different from traditional vision Transformers~\cite{dosovitskiy2020image,liu2021swin,cao2021video}, which crop image patches and form a set of spatial visual tokens, our tokens are based on different frequency bands. In a nutshell, we generate $N$ blocks for each spectral frame $D^t_{LR}$, and each block has  DCT-based frequency tokens $\tau$ of the number of $F$. The total number of frequency tokens is $T\times N \times F$. Figure \ref{fig:FTVSR} (a) presents more details about the whole tokenization process.

\begin{figure}[!t]
\centering
\includegraphics[width=\columnwidth]{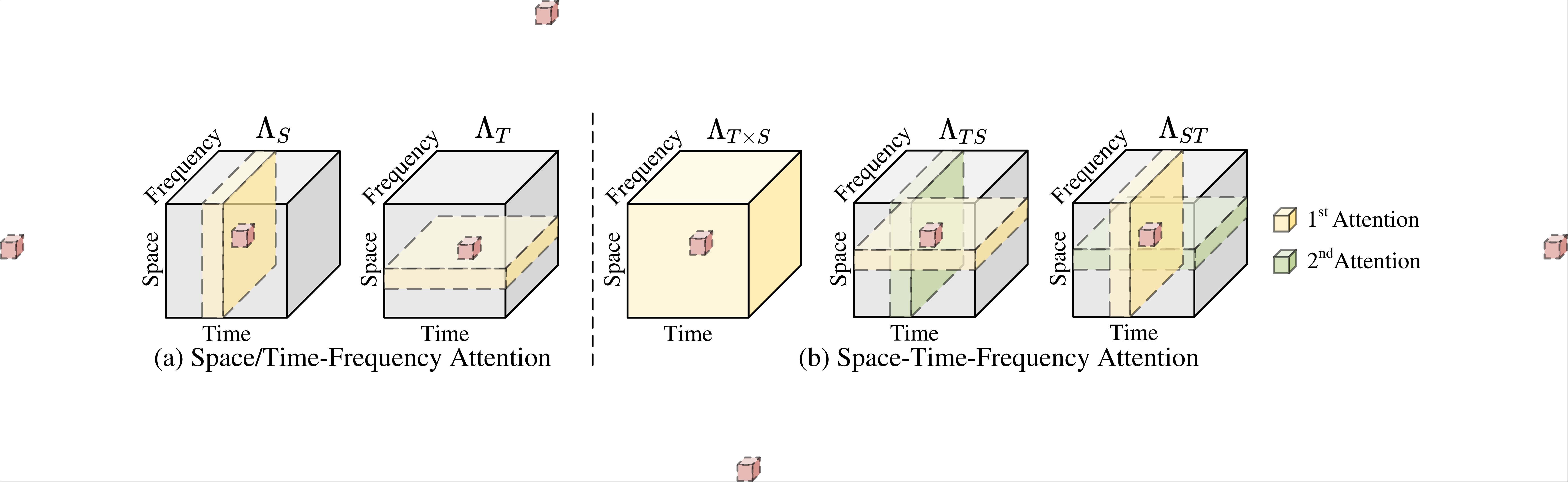}
\caption{The visualization of (a) space/time-frequency attention, and (b) space-time-frequency attention. The red cube denotes the query token. The yellow area and green area represent the candidate area for computing attention with query following the order of yellow first and then green
}
\label{fig:freqAttention}
\end{figure}

\subsection{Frequency-based Attention}
\label{frequecny_transformer}
The inputs of frequency transformer are DCT-based visual tokens, which have been generated in Section \ref{dct_tokens}. To better take advantage of temporal information for VSR, the query tokens $\mathcal{Q}$ are extracted from spectral map $D^T_{LR}$. Keys $\mathcal{K}$ and values $\mathcal{V}$ are extracted from spectral maps $\{D^t_{LR}, t\in [1, T-1]\}$. For a target frame $D^T_{LR}$, the query, key, and value sets are denoted as:
\begin{equation}
\begin{aligned}
    \mathcal{Q} & = \{ \tau^q_{(T,i,f)}, i\in [1,N], f\in [1,F] \}, \\
    \mathcal{K} & = \{ \tau_{(t,i,f)}^k, t\in [1,T-1],i\in [1,N], f\in [1,F]\}, \\
    \mathcal{V} & = \{ \tau_{(t,i,f)}^v, t\in [1,T-1],i\in [1,N], f\in [1,F]\},
\end{aligned}
\end{equation}
where $\tau_{(T,i,f)}^q$, $\tau_{(t,i,f)}^k$, and $\tau_{(t,i,f)}^v$ represent the query, key, and value tokens, respectively. Each token is extracted from spectral maps among time, space, and frequency dimensions according to needs of computing different kinds of frequency attention, which will be discussed as follows.

\subsubsection{Frequency Attention}

The frequency attention aims to capture the relationship between different frequency bands.
Given a query token $\tau^q_{f}$ at the $f^{th}$ frequency, the uniform formulation of frequency attention (denoted as $\Lambda$) is:
\begin{equation}
\label{eq_fa}
    \Lambda (\tau^q_{f}, \tau_{\hat{f}}^k, \tau_{\hat{f}}^v) = \text{SM}(\frac{\tau_{f}^q \cdot \tau_{\hat{f}}^k}{\sqrt{d^k}})\tau_{\hat{f}}^v, \hat{f} \in [1,F],
\end{equation}
where $\text{SM}$ represents the softmax activation function and $d^k$ denotes the normalization factor. Note that there is a feed forward network (FFN) after frequency attention, which is omitted in this paper.
However, computing frequency attention on whole spectral maps is impractical since the feature size of spectral map is different during the process of training and inference. Therefore, we adopt the way of computing frequency attention on spatial blocks of spectral maps. To explore different frequency attention mechanisms combined with time or space attention. As shown in Figure \ref{fig:freqAttention}, we propose space-frequency attention, time-frequency attention, and time-space-frequency attention.

\subsubsection{Space/Time-Frequency Attention}
Space-Frequency (SF) attention computes the frequency attention weights between spatial blocks. The visualization of SF is shown in Figure \ref{fig:freqAttention} (a). 
For a query token $\tau^q_{(i,f)}$ at the $f^{th}$ frequency in the $i^{th}$ block, the SF attention is $\Lambda_S(\tau^q_{(i,f)}, \tau^k_{(\hat{i},\hat{f})}, \tau^v_{(\hat{i},\hat{f})}), \hat{i}\in [1,N], \hat{f}\in [1,F]$, which computes the frequency attention as Equation \ref{eq_fa} in spatial dimension. The inputs of $\Lambda_S$ are space-frequency tokens $\tau_{(i,f)}$. Since the tokens are extracted from both space and frequency dimensions, $N\times F$ tokens are generated for SF attention. 

The Time-Frequency (TF) attention is computed on the blocks with the same spatial position from different video frames. The visualization of TF attention is shown in Figure \ref{fig:freqAttention} (a). 
Given a query token $\tau^q_{(t,f)}$, the TF attention is $\Lambda_T(\tau^q_{(t,f)}, \tau^k_{(\hat{t}, \hat{f})}, \tau^v_{(\hat{t}, \hat{f})}), \hat{t}\in [1,T-1],\hat{f}\in [1,F]$, which computes frequency attention as Equation \ref{eq_fa} in temporal dimension.
The inputs of $\Lambda_T$ are time-frequency tokens $\tau_{(t,f)}$. Since the tokens are extracted from both time and frequency dimensions, $T\times F$ tokens are generated for $TF$ attention.

\subsubsection{Time-Space-Frequency Attention}
Both the temporal and spatial information are important for compressed VSR. To further explore the frequency attention in both spatial and temporal dimensions, we propose Time-Space-Frequency (TSF) attention.
TSF are the combinations of SF and TF attention. It can be divided into two types: joint SF and TF attention, divided SF and TF attention. The visualizations of TSF are shown in Figure \ref{fig:freqAttention} (b). Given a query token $\tau^q_{(t,i,f)}$, joint TSF attention is $\Lambda_{T\times S}(\tau^q_{(t,i,f)}, \tau^k_{(\hat{t},\hat{i},\hat{f})}, \tau^v_{(\hat{t},\hat{i},\hat{f})}), $ $\hat{t}\in [1,T-1], \hat{i} \in [1,N],\hat{f}\in [1,F]$, which computes the frequency attention as Equation \ref{eq_fa} in both spatial and temporal dimensions. 
The inputs of joint TSF attention are time-space-frequency tokens $\tau_{(t,i,f)}$. Since the tokens are extracted among time, space, and frequency dimensions, $T\times N \times F$ tokens are generated for joint TSF.

For divided TSF attention, two types of TSF are designed according to the order of computing $TF$ and $SF$ attention. One of them can be formulated as: 
\begin{equation}
\label{eq_ST}
\begin{aligned}
      &\Lambda_{ST}(\tau^q_{(t,i,f)}, \tau^k_{(\hat{t},\hat{i},\hat{f})}, \tau^v_{(\hat{t},\hat{i},\hat{f})}) = \Lambda_{T}(\hat{\tau}^q_{(t,f)}, \tau_{(\hat{t},\hat{f})}, \tau_{(\hat{t},\hat{f})}), \\
      where ~\hat{\tau} &= \Lambda_{S}(\tau^q_{(i,f)}, \tau^k_{(\hat{i},\hat{f})}, \tau^k_{(\hat{i},\hat{f})}), \hat{t}\in [1,T-1], \hat{i}\in [1,N], \hat{f}\in [1,F].
\end{aligned}
\end{equation}
The divided TSF attention $\Lambda_{ST}$ represents the attention that computes space-frequency attention $\Lambda_S$ firstly, then computes time-frequency attention $\Lambda_T$. 

In our experiments, $\Lambda_{ST}$ performs best for frequency transformer. This is because in compressed VSR, degraded frames should be first restored by the space-frequency attention then the recovered textures could be used to benefit temporal learning in the time-frequency attention. 
The other one can be formulated as $\Lambda_{TS}(\tau^q_{(t,i,f)}, \tau^k_{(\hat{t},\hat{i},\hat{f})}, \tau^v_{(\hat{t},\hat{i},\hat{f})})$.
The divided TSF attention $\Lambda_{TS}$ represents the attention that computes time-frequency attention $\Lambda_{T}$ firstly, then computes space-frequency attention $\Lambda_{S}$. The computing process of $\Lambda_{TS}$ is similar with Equation \ref{eq_ST}.

\subsection{Frequency Transformer}
To recover HR sequences, we use the similar recurrent structure as TTVSR~\cite{liu2022learning}. Each HR frame is restored from its LR counterparts and a propagation hidden state $H$. Given a LR frame $I^T_{LR}$, the SR frame can be restored as:
\begin{equation}
\begin{aligned}
    I^T_{SR} & = \text{rDCT}(T_{freq}(\mathcal{Q}, \mathcal{K}, \mathcal{V})) \\
    & = \text{rDCT}(\Gamma(A_{freq}(\mathcal{Q}, \mathcal{K}, \mathcal{V}), D_{LR}^T)+ D_{LR}^T),
\end{aligned}
\end{equation}
where $T_{freq}$ represents the Frequency Transformer. $A_{freq}$ represents the frequency attention used in $T_{freq}$ and $A_{freq} \in \{\Lambda_S, \Lambda_T, \Lambda_{T\times S}, \Lambda_{TS}, \Lambda_{ST}\}$. $\Gamma$ represents the fusion operation which concatenates the outputs of $A_{freq}$ and $D^T_{LR}$, then reduces the dimensions of the concatenated features by Linear layer.

For example, a frequency Transformer formed by divided TSF attention $\Lambda_{SF}$ is shown in Figure \ref{fig:FTVSR} (b). The output $P^T$ of TSF can be formulated as:
\begin{equation}
\begin{aligned}
P^T &~= \Lambda_T(R_T, \Hat{H}^T, \Hat{H}^T),\\
where ~R^T &~= \Lambda_S(\mathcal{Q}_S, \mathcal{K}_S,  \mathcal{V}_S), \Hat{H}^T = W(H^{T-1}, O^T).
\end{aligned}
\end{equation}
$\Hat{H}^T$ represents the hidden states warped from past frames $H^{T-1}$ according to flow $O^T$. $W$ represents the flow warp operation as \cite{chan2021basicvsr}. $H^T$ is updated by the output $P^T$ of TSF attention and the DCT-based features $D^T_{LR}$. $\mathcal{Q}_S$ are extracted from upsampled $I^T_{LR}$ by Bicubic upsampling while $\mathcal{K}$ and $\mathcal{V}$ are extracted from upsampled $I^T_{LR}$ by a upsample neural network. The difference between upsample operations brings the location guidance of the hard-to-recover parts, which should pay more attention to it. $\mathcal{Q}_T$ is the temporal-frequency query tokens for $\Lambda_T$. The output $P^T$ of TSF attention $\Lambda_{SF}$ is used to recover SR frames, which can be formulated as:
\begin{equation}
\label{eq_final_ftvsr}
  I^T_{SR}  = \text{rDCT}(\Gamma(P^T,D_{LR}^T) + D_{LR}^T).
\end{equation}
More details about the network structure of our proposed Frequency-Transformer can be found in the supplementary material.

We follow the previous works~\cite{chan2021basicvsr,li2021comisr}, using Charbonnier penalty loss~\cite{lai2017deep}, which is applied on each video frames. The total loss $\mathcal{L}$ is the average of frames, 
\begin{equation}
    \mathcal{L} = \frac{1}{T}\sum_{t=1}^T\sqrt{||I^t_{HR} - I^t_{SR}||^2 + \epsilon^2},
\end{equation}
where $\epsilon$ is a constant value and $\epsilon =1e-3$.

\section{Experiments}

\subsection{Implementation Details}
During training, the Cosine Annealing scheme and Adam optimizer with $\beta_1=0.9$ and $\beta_2=0.99$ are used. The initial learning rate of FTVSR is $2\times 10^4$. The batch size is 8 videos. The training frame length is 40 for final results and 10 for ablation study. The input patch size is $64\times 64$ and the SR scale is $4\times$. Data augmentations include random horizontal flips, vertical flips, and rotations. We train FTVSR with 400k iterations for the final model and 100k iterations for quick ablation study. All ablation study are based on the backbone of BasicVSR~\cite{chan2021basicvsr} and final model is based on the backbone of TTVSR~\cite{liu2022learning} for better results.

Unless otherwise stated, FTVSR is trained with a ratio of $50\%$ uncompressed videos and $50\%$ compressed videos. The compressed videos are uniformly sampled from different compression rates. During inference, we pad the input images with the edge values to keep they can be transformed into spectral maps by DCT and remove the padding after transforming spectral maps into pixel images by rDCT. We crop images into $4\times 4$ patches for inference since the limitation of CUDA memory.

\subsection{Datasets and Evaluation Metrics}
\subsubsection{Datasets}
Following the previous works~\cite{cao2021video,chan2021basicvsr}, we use REDS~\cite{nah2019ntire} and Vimeo-90K~\cite{xue2019video} for training. The REDS dataset contains 270 videos and each video in has 100 frames with a resolution of $1280\times 720$. For a fair comparison, four sequences as previous works~\cite{chan2021basicvsr,cao2021video,wang2019edvr,li2020mucan} for testing, called REDS4. The Vimeo-90K contains 64,612 sequences for training. Each video contains 7 frames with a resolution of $448\times 256$. Same as previous works~\cite{li2021comisr,chan2021basicvsr}, the testing set of Vimeo-90K is Vid4, which contains four videos. Each video includes $30$ to $50$ frames.
\subsubsection{Evaluation Metrics}
We use the same metrics peak signal-to-noise ratio (PSNR) and structural similarity index (SSIM)~\cite{wang2004image} as previous works~\cite{chan2021basicvsr,cao2021video,wang2019edvr,li2020mucan} in our evaluation. In addition, for compression videos, we use the most common setting for H.264 codec at different compression rates (different CRF values). Following previous COMISR~\cite{li2021comisr}, we use CRF of 15, 25, and 35 to generate compressed videos. Detailed command for video compression can be found in the supplementary material. We then evaluate FTVSR and report the PSNR and SSIM on these compressed videos with these CRF values.

\subsection{Comparison with State-of-the-art Methods}

\begin{table*}[t]
  \caption{Quantitative comparison on the \textbf{compressed} videos of REDS4~\cite{nah2019ntire} for $4\times$ VSR. Each entry shows the PSNR$\uparrow$/SSIM$\uparrow$ on RGB channels as \cite{chan2021basicvsr,li2021comisr}. \textcolor{red}{Red} indicates the best and \textcolor{blue}{{blue}} indicates the second best performance (Best viewed in color)}
  \centering
  \renewcommand\arraystretch{1.2}
  \renewcommand\tabcolsep{3pt}
  \resizebox{\columnwidth}{!}{
  \begin{tabular}{ l| c | c | c | c | c | c | c}
    \hline
    
    \hline
    \multirow{2}{*}{Method} &\multicolumn{4}{c|}{Per clip with Compression CRF25} &\multicolumn{3}{c}{Average of clips with Compression}\\
    \cline{2-8}
      &  Clip\_000 &  Clip\_011 &  Clip\_015 &  Clip\_020 & CRF15 & CRF25 & CRF35 \\
    \hline
    DUF~\cite{jo2018deep} & 23.46/0.622 & 24.02/0.686 & 25.76/0.773 & 23.54/0.689 & 25.61/0.775 & 24.19/0.692 & 22.17/0.588 \\
    FRVSR~\cite{sajjadi2018frame} & 24.25/0.631 & 25.65/0.687 & 28.17/0.770 & 24.79/0.694 & 27.61/0.784 & 25.72/0.696 & 23.22/0.579 \\
    EDVR~\cite{wang2019edvr} & 24.38/0.629 & 26.01/0.702 & 28.30/0.783 & 25.21/0.708 & 28.72/0.805 & 25.98/0.706 & 23.36/0.600 \\
    TecoGan~\cite{chu2020learning} & 24.01/0.624 & 25.39/0.682 & 27.95/0.768 & 24.48/0.686 & 26.93/0.768 & 25.46/0.690 & 22.95/0.589 \\
    RSDN~\cite{isobe2020video} & 24.04/0.602 & 25.40/0.673 & 27.93/0.766 & 24.54/0.676 & 27.66/0.768 & 25.48/0.679 & 23.03/0.579 \\
    MuCAN~\cite{li2020mucan} & 24.39/0.628 & 26.02/0.702 & 28.25/0.781 & 25.17/0.707 & 28.67/0.804 & 25.96/0.705 & 23.55/0.600 \\
    BasicVSR~\cite{chan2021basicvsr} &24.37/0.628 & 26.01/0.702 & 28.13/0.777&25.21/0.709 & 29.05/0.814 & 25.93/0.704 & 23.22/0.596\\
    IconVSR ~\cite{chan2021basicvsr} &24.35/0.627 & 26.00/0.702 & 28.16/0.777& 25.22/0.709&\color{blue}{{29.10/0.816}} &25.93/0.704 & 23.22/0.596\\
    COMISR~\cite{li2021comisr} & \color{blue}{{24.76/0.660}} & \color{blue}{{26.54/0.722}} & \color{blue}{{29.14/0.805}} & \color{blue}{{25.44/0.724}} &
    28.40/0.809 & 
    \color{blue}{{26.47/0.728}} & \color{blue}{{23.56/0.599}}\\
    \hline
    \textbf{FTVSR} & 
    \color{red}{26.06/0.703}& 
    \color{red}{28.71/0.779} & 
    \color{red}{30.17/0.839} & 
    \color{red}{27.26/0.782} & 
    \color{red}{30.51/0.853} & 
    \color{red}{28.05/0.776} & 
    \color{red}{24.82/0.657}\\
    \hline
    
    \hline
  \end{tabular}
  }
  \label{tab_compress_reds}
\end{table*}

\subsubsection{Evaluation on Compressed Videos}
We compare FTVSR with other state-of-the-art methods on REDS~\cite{nah2019ntire} and Vid4~\cite{xue2019video} datasets. Following the compressed settings as COMISR~\cite{li2021comisr}, we compress the videos with several compression rates (CRF15, CRF25, CRF35) and evaluate on the compressed videos in PSNR and SSIM. 

For REDS~\cite{nah2019ntire} dataset, the results on compressed videos are shown in Table \ref{tab_compress_reds} and results of other methods are cited from \cite{li2021comisr}. For BasicVSR and IconVSR, we finetune them on the compressed videos as the same training settings of \cite{li2021comisr}. 
Although recent BasicVSR~\cite{chan2021basicvsr} and IconVSR~\cite{chan2021basicvsr} achieve state-of-the-art results on uncompressed videos, they perform not well on the compressed videos. 
For example, BasicVSR achieves 25.93dB and 23.22dB in PSNR of compression CRF25 and CRF35.
Besides, IconVSR, which performs better than BasicVSR on uncompressed videos, but just obtain 25.93dB and 23.22dB in PSNR of compression CRF25 and CRF35 same as the BasicVSR.
This phenomenon indicates that only increases the model capacity has less effect on compression problems.

COMISR~\cite{li2021comisr} alleviates the compression problem to some extent by its special designs for compression, but the gains are small (e.g., 26.47dB and 23.56dB in PSNR with a compression rate of CRF25 and CRF35). 
However, FTVSR achieves 30.51, 28.05dB, and 24.82dB in PSNR on compressed videos with a compression rate of CRF15, CRF25, and CRF35, respectively. FTVSR outperforms SOTA COMISR by 1.6dB on the compressed videos in CRF25. The results show that FTVSR has strong capabilities on compression problems.

\begin{table*}[t]
  \caption{Quantitative comparison on the \textbf{compressed} video of Vid4~\cite{xue2019video} for $4\times$ VSR. Following previous works~\cite{chan2021basicvsr,li2021comisr}, each entry shows the PSNR$\uparrow$/SSIM$\uparrow$ on Y-channel. \textcolor{red}{Red} and \textcolor{blue}{{blue}} indicates the best and second best performances (Best viewed in color)}
  \centering
  \renewcommand\tabcolsep{3pt}
  \resizebox{\columnwidth}{!}{
  \begin{tabular}{ l| c | c | c | c | c | c | c}
    \hline
    
    \hline
    \multirow{2}{*}{Method}& \multicolumn{4}{c|}{Per clip with Compression CRF25} &\multicolumn{3}{c}{Average of clips with Compression}\\
    \cline{2-8}
      &  calendar &  city & foliage & walk & CRF15 & CRF25 & CRF35 \\
    \hline
    DUF~\cite{jo2018deep} & 21.16/0.634 & 23.78/0.632 & 22.97/0.603 & 24.33/0.771 & 24.40/0.773 & 23.06/0.660 & 21.27/0.515 \\
    FRVSR~\cite{sajjadi2018frame}& 21.55/0.631 & 25.40/0.575 & 24.11/0.625 & 26.21/0.764 &26.01/0.766 & 24.33/0.655 & 22.05/0.482 \\
    EDVR~\cite{wang2019edvr} & 21.69/0.648 & 25.51/0.626 & 24.01/0.606 & 26.72/0.786 & 26.34/0.771 & 24.45/0.667 & 22.31/0.534 \\
    TecoGan~\cite{chu2020learning} & 21.34/0.624 & 25.26/0.561 & 23.50/0.592 & 25.73/0.756 & 25.25/0.741 & 23.94/0.639 & 21.99/0.479 \\
    RSDN~\cite{isobe2020video} & 21.72/0.650 & 25.28/0.615 & 23.69/0.591 & 25.57/0.747 & 26.58/0.781 & 24.06/0.650 & 21.29/0.483 \\
    MuCAN~\cite{li2020mucan} & 21.60/0.643 & 25.38/0.620 & 23.93/0.599 & 26.43/0.782 & 25.85/0.753 & 24.34/0.661 & 22.26/0.531 \\

    BasicVSR~\cite{chan2021basicvsr} & 21.64/0.641 & 25.45/0.620 &23.79/0.586 &26.26/0.774 &26.56/0.780 &24.28/0.656 & 21.97/0.509 \\
    IconVSR ~\cite{chan2021basicvsr} & 21.67/0.644 &25.46/0.621 &23.83/0.588 & 26.26/0.774& \color{blue}{26.65/0.782} &24.31/0.657 &21.97/0.509 \\
    COMISR~\cite{li2021comisr} & 
    \color{blue}{22.81/0.695} & 
    \color{blue}{25.94/0.640} & 
    \color{blue}{24.66/0.656} & 
    \color{blue}{26.95/0.799} & 
    26.43/0.791 & 
    \color{blue}{24.97/0.701} & 
    \color{blue}{22.35/0.509} \\
    \hline
    \textbf{FTVSR} & 
    \color{red}{22.97/0.720}&
    \color{red}{26.29/0.670} & 
    \color{red}{24.94/0.664} & 
    \color{red}{27.30/0.816} & 
    \color{red}{27.40/0.811} & 
    \color{red}{25.38/0.706} & 
    \color{red}{22.61/0.540}\\
    \hline
    
    \hline
  \end{tabular}
  }
  \label{tab_com_vid4}
\end{table*}

\begin{table*}[!t]
  \caption{Evaluation on the \textbf{uncompressed} videos of REDS4~\cite{nah2019ntire} and Vid4~\cite{xue2019video} for $4\times$ VSR. Each entry shows the PSNR$\uparrow$/SSIM$\uparrow$. $*$ represents the FTVSR is trained on only uncompressed videos. $\dagger$ represents FTVSR is trained on both compressed and uncompressed videos, which is a more difficult setting. All other methods are trained on uncompressed videos and evaluated on uncompressed videos}
  \centering
  \renewcommand\arraystretch{1.2}
  \renewcommand\tabcolsep{2pt}
  \resizebox{\columnwidth}{!}{
  \begin{tabular}{ l|c|c|c|c|c|c|c|c}
    \hline
    
    \hline
    Datasets  & TOFlow\cite{xue2019video} & DUF\cite{jo2018deep}  & EDVR\cite{wang2019edvr} & COMISR\cite{li2021comisr} & BasicVSR\cite{chan2021basicvsr} & IconVSR\cite{chan2021basicvsr} & \textbf{FTVSR}$^*$ & \textbf{FTVSR$^\dagger$}\\
    \hline
    REDS4 & 27.98/0.799 & 28.63/0.825 & 31.09/0.880 & 29.68/0.868 & 31.42/0.890 & 31.67/0.895 & 31.82/0.896 & 31.74/0.895 \\
    \hline
    Vid4 & 25.85/0.766 & 27.38/0.832 & 27.85/0.850 & 27.31/0.840 & 27.96/0.855 & 28.04/0.857 & 28.31/0.860 & 28.06/0.856\\
    \hline
    
    \hline
  \end{tabular}
  }
  \label{tab_reds_unc}
\end{table*}

For Vid4~\cite{xue2019video} dataset, the results on compressed videos are shown in Table \ref{tab_com_vid4}. The results of BasicVSR and IconVSR are obtained by finetuning on compressed videos as \cite{li2021comisr}. 
For a fair comparison, we also adopt the same compression settings as COMISR~\cite{li2021comisr}. On compressed videos with a compression rate of CRF 15, 25, and 35, FTVSR achieves 27.40dB, 25.38dB, and 22.61dB in PSNR, respectively. FTVSR outperforms other methods. These results demonstrate the huge potential of FTVSR on the task of compressed VSR.

\begin{figure}[!t]
\centering
\includegraphics[width=0.95\columnwidth]{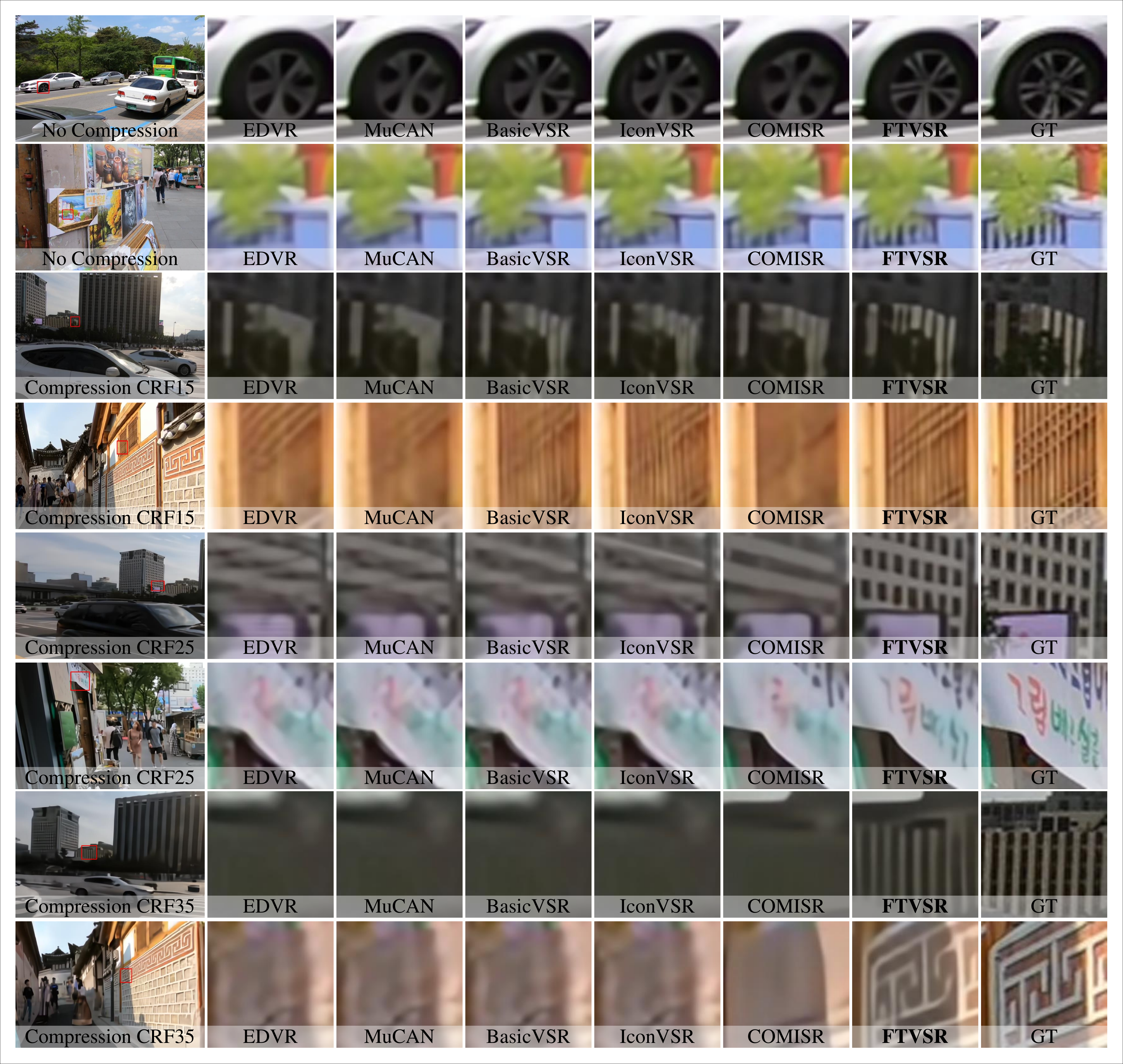}
\vspace{-0.5cm}

\caption{Visualization results of our FTVSR and other VSR methods on the uncompressed videos and compressed videos with compression rates of CRF 15, 25, and 35}
\label{fig:case}
\end{figure}

We also visualize the results of FTVSR and SOTA methods on compressed videos. As shown in Figure \ref{fig:case}, FTVSR performs well on both compressed and uncompressed video. Especially on the compressed video with CRF25 and CRF35, the visual quality of FTVSR is superior to other methods. An interesting phenomenon is that COMISR performs better than BasicVSR and IconVSR on compressed videos, but poorly on uncompressed videos.  However, our FTVSR performs well on both compressed videos and uncompressed videos as shown in Figure \ref{fig:case}. Especially on the cases with a compression rate of CRF35, BasicVSR, IconVSR, and COMISR are failed to recover the texture, but FTVSR still performs well on these cases. It's because frequency attention enables low-frequency information to guide the generation of high-frequency textures.

\begin{table}[!t]
  \caption{Comparison of parameters, FLOPs and PSNR$\uparrow$/SSIM$\uparrow$ on the compressed videos with CRF25. FLOPs is computed on one LR frame with the size of $180 \times320$ and $\times 4$ upsampling on the REDS4 dataset}
  \centering
    \renewcommand\tabcolsep{4pt}
  \resizebox{\columnwidth}{!}{
  \begin{tabular}{ c|c|c|c|c|c|c|c}
  \hline
  
  \hline
  Methods & DUF\cite{jo2018deep} & EDVR\cite{wang2019edvr} & MuCAN\cite{li2020mucan} & BasicVSR\cite{chan2021basicvsr} & IconVSR\cite{chan2021basicvsr} & COMISR\cite{li2021comisr}& FTVSR\\
  \hline
  Params(M) & 5.8 & 20.6 & 13.6 & 6.3 & 8.7 & 6.2 & 10.8 \\
  FLOPs(T) & 2.34  &  2.95 & $>$1.07 & 0.33 & 0.51 &0.36 & 0.76\\
  \hline
  PSNR/SSIM & 24.19/0.692 & 25.98/0.706 & 25.96/0.705 & 25.93/0.704 & 25.93/0.704 & 26.47/0.728 & 27.28/0.763\\

    \hline 
    
    \hline
  \end{tabular}
  }
  \label{tab_params}
\end{table}

\begin{figure}[]
\centering
\includegraphics[width=0.95\columnwidth]{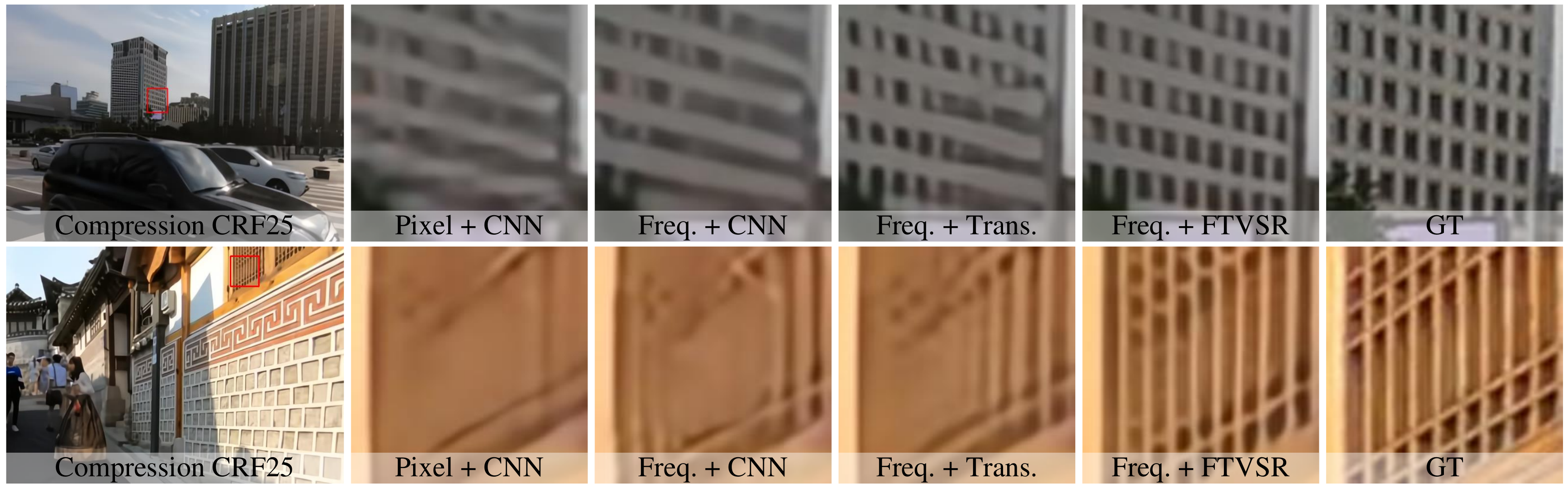}
\vspace{-0.3cm}

\caption{Visualization results of ``Pixel + CNN", ``Frequency + CNN", ``Frequency + Transformer", and ``Frequency + FTVSR" in Table \ref{tab_freq}}
\label{fig:ablation}
\end{figure}

\subsubsection{Evaluation on Uncompressed Videos}
To study the potential of FTVSR, we also evaluate FTVSR on the uncompressed videos of the REDS and Vid4 datasets, respectively. 
For a fair comparison, we compare with SOTA methods in two settings: 
1) FTVSR$^*$, training only on uncompressed videos, and testing on uncompressed videos.
2) FTVSR$^\dagger$, training on both compressed and uncompressed videos, and testing on uncompressed videos, which is a more difficult setting for evaluating on uncompressed videos since the compressed data brings more noises for VSR model. 
As shown in Table \ref{tab_reds_unc}, all results are evaluated on uncompressed videos. The results of other methods are all obtained from their paper and their model is trained on only uncompressed videos as setting 1. 
For the setting 1 which is fair for our method, FTVSR$^*$ outperforms SOTA IconVSR~\cite{chan2021basicvsr} in PSNR on both REDS4 and Vid4 datasets. 
Moreover, for setting 2, FTVSR$^\dagger$ achieves 31.74dB in PSNR on REDS4 dataset, which outperforms IconVSR~\cite{chan2021basicvsr} although IconVSR is trained on full clean uncompressed videos. 
Besides, FTVSR$^\dagger$ obtain comparable results on Vid4 dataset. Compared with COMISR~\cite{li2021comisr} that performs well on compressed videos while unsatisfactory on uncompressed videos, FTVSR performs well on both compressed videos and uncompressed videos, even in the difficult setting of that the model is trained on both compressed videos and uncompressed videos.

\subsubsection{Comparison of Parameters and FLOPs}
The comparisons of parameters and FLOPs are shown in Table \ref{tab_params}. The FLOPs are computed with the input of LR frame size $180\times 320$ and conducting $4\times$ upsampling VSR task. Based on BasicVSR, FTVSR outperforms other methods with the comparable parameters and FLOPs. FTVSR adopt a similar architecture as BasicVSR and process HR frames in the frequency domain. The FLOPs are comparable with BasicVSR since the DCT operation reduces the computational costs of FTVSR.

\begin{table*}[!t]
  \caption{Ablation study of FTVSR (PSNR$\uparrow$/SSIM$\uparrow$) on the REDS4 dataset}
  \centering
    \renewcommand\arraystretch{1.2}
  \renewcommand\tabcolsep{3pt}
  \resizebox{\columnwidth}{!}{
  \begin{tabular}{ l | c | c | c | c | c | c | c}
    \hline
    
    \hline
    \multirow{2}{*}{Domain + Backbone} &\multicolumn{4}{c|}{Per clip with Compression CRF25} &\multicolumn{3}{c}{Average of clips with Compression}\\
    \cline{2-8}
     &  Clip\_000 &  Clip\_011 &  Clip\_015 &  Clip\_020 & CRF15 & CRF25 & CRF35 \\
    \hline
    Pixel + CNN& 24.37/0.628 & 26.01/0.702 & 28.13/0.777 & 25.21/0.709 & 29.05/0.814 & 25.93/0.704 & 23.22/0.596\\
    Frequency + CNN& 24.98/0.666& 27.11/0.746 & 29.36/0.818 & 26.05/0.751 & 29.20/0.825 & 26.87/0.745 & 23.83/0.629 \\
    
    Frequency + Transformer & 25.20/0.684 & 27.53/0.763 & 29.47/0.828 & 26.33/0.766 & 29.51/0.837 & 27.15/0.759 & 24.03/0.644 \\
    Frequency + FTVSR & 25.26/0.609 & 27.75/0.766 & 29.62/0.831 & 26.47/0.772 & 29.70/0.843 & 27.28/0.763 & 24.22/0.646\\
    \hline
    
    \hline
  \end{tabular}
  }
  \label{tab_freq}
\end{table*}

\begin{table*}[!t]
  \caption{Comparisons of different types of frequency attention on the compressed videos of REDS with compression rates of CRF 15, 25, and 35. All the methods in this table are in the frequency domain. ``Base" represents the traditional attention without frequency attention mechanism. Each entry shows PSNR$\uparrow$/SSIM$\uparrow$}
  \centering
  \renewcommand\arraystretch{1.2}
  \renewcommand\tabcolsep{8pt}
  \resizebox{\columnwidth}{!}{
  \begin{tabular}{ c|c|c|c|c|c|c}
    \hline

    \hline
     Attention & Base & $\Lambda_S$ &  $\Lambda_T$ &  $\Lambda_{T\times S}$ &  $\Lambda_{TS}$ &  $\Lambda_{ST}$ \\
     \hline 
     CRF15& 29.51/0.837 & 29.63/0.840 & 29.60/0.840 &29.61/0.839 & 29.65/0.841 & \textbf{29.70/0.843}\\
     CRF25& 27.15/0.759 & 27.23/0.761 & 27.11/0.760 & 27.22/0.760& 27.24/0.762 &\textbf{27.28/0.763}  \\
     CRF35& 24.03/0.644 & 24.12/0.646 & 24.05/0.641 &24.11/0.644 & 24.12/0.645 &\textbf{24.22/0.646} \\

    \hline
    
    \hline
  \end{tabular}
  }
  \label{tab_abattention}
\end{table*}

\subsection{Ablation Study}

To evaluate the effectiveness of FTVSR, we conduct the ablation study on the REDS4 dataset. As shown in Table \ref{tab_freq}, We use BasicVSR~\cite{chan2021basicvsr} as baseline, which learns in pixel domain and achieves 29.05dB, 25.93dB, 23.22dB in PSNR on the compressed videos with compression CRF15, CRF25, and CRF35, respectively. 
The performances are poor compared with its 31.42dB on uncompressed videos. 
Then, we transfer images into the frequency domain, which obtains a relative gain of 0.94dB in PSNR on the compressed videos with CRF25. 
In the frequency domain, a transformer-based model without frequency attention achieves 27.15dB in PSNR on the compressed videos with CRF25, which shows that the attention mechanism is beneficial for frequency learning.
Replacing the basic transformer by our FTVSR, FTVSR achieves 27.28dB in PSNR on the compressed videos with CRF25, which shows that the frequency attention is better than traditional attention in the frequency domain. As shown in Figure \ref{fig:ablation}, FTVSR achieves better visualization results than others.

To evaluate the effectiveness of different frequency attention  introduced in Section \ref{frequecny_transformer}, we conduct the ablation study on the REDS dataset. As shown in Table \ref{tab_abattention},``Base" represents a traditional transformer which computes spatial attention in frequency domain.
The results of base attention are lower than frequency attention.
For the different frequency attentions, space-frequency attention, time-frequency attention, joint time-space-frequency attention and divided time-space-frequency attention ($\{\Lambda_S, \Lambda_T, \Lambda_{T\times S}, \Lambda_{TS}, \Lambda_{ST}\}$), the results in Table \ref{tab_abattention} show that the divided frequency attention ($\Lambda_{ST}$) with an order of space first and time later is better. This is because in compressed VSR, degraded frames should be first restored by the space-frequency attention then the recovered textures could be used to benefit temporal learning in the time-frequency attention.

\section{Conclusions}
In this paper, we propose a novel spatiotemporal Frequency-Transformer for compressed Video Super-Resolution (FTVSR). To handle the compression issues, we transform compressed video frames into frequency domain and design frequency-based attention to enable the feature fusions across multiple frequency bands. The frequency-based tokenization and frequency attention mechanism enables low-frequency information to guide the generation of high-frequency textures. 
To utilize spatial and temporal information, we further explore the different types of frequency attention combined with space and time attentions.
Experiments on two widely-used VSR datasets show that the proposed FTVSR significantly outperforms previous works and achieves new SOTA results.

\section*{Acknowledgments}
This work was supported by the Scientific and Technological Innovation of Shunde Graduate School of University of Science and Technology Beijing (No. BK20AE004 and No.BK19CE017).



\clearpage
%
%
\bibliographystyle{splncs04}
\bibliography{egbib}

\newpage
\begin{appendix}
\section*{Supplementary Material}

In this supplementary material, we introduce the algorithm details in Section \ref{sec:alg}. More implemental details are shown in Section \ref{sec:implemental}. The influence of compression is discussed in Section \ref{sec:compress}. More visualization results and failure cases are shown in Section \ref{sec:vis}. The limitations of FTVSR are discussed in Section \ref{sec:vis}.

\section{Algorithm Details}
\label{sec:alg}

The algorithm details of FTVSR with divided time-space-frequency attention $\Lambda_{ST}$ are shown in Algorithm \ref{alg}. In Algorithm \ref{alg}, the upsampling network $\varphi(\cdot)$ and the flow network $\phi(\cdot)$ is same as TTVSR~\cite{liu2022learning}.  We follow COMISR~\cite{li2021comisr}, BasicVSR~\cite{chan2021basicvsr} and TTVSR~\cite{liu2022learning} to use bidirectional propagation scheme, which includes forward and backward propagation. For clarity, only forward propagation is shown in Algorithm \ref{alg}.

\begin{algorithm}[h]\small
  \caption{FTVSR with divided time-space-frequency attention $\Lambda_{ST}$} 
  \label{alg}
  \begin{algorithmic}[1]
    \Require
      $\mathbf{I}_{LR}$: $\{I_{LR}^{t}, t \in [1,T]\}$;
      $T$: the length of sequence;
      $N$: the block numbers of each frame;
      $F$: the frequency numbers.
      $H_{init}$ initialization by zero.
      $U(\cdot)$: Bicubic upsampling.
      $\varphi(\cdot)$: upsampling network.
      $\phi(\cdot)$: flow estimation.
      $\text{DCT}(\cdot)$: Discrete Cosine Transform.
      $\text{rDCT}(\cdot)$: inverse Discrete Cosine Transform.
      $W(\cdot)$: flow warp.
      $\Lambda_{S}(\cdot)$: space-frequency attention.
      $\Lambda_{T}(\cdot)$: time-frequency attention.
      $\Gamma$: fusion layer.
    \Ensure
      $\mathbf{I}_{SR}$: $\{I_{SR}^{t}, t \in [1,T]\}$;\\
      $H=\{H_{init}\}$;
      \For{$t = 1$; $t<=T$; $t++$}
          \State $O^{t} = \phi(I_{LR}^t, I_{LR}^{t-1})$;
          \State $\Hat{H}^t = W(H^{t-1}, O^t)$;
          \State $\mathcal{Q} = \text{DCT}(U(I^t_{LR}))=\{\tau^q_{(t,i,f)}, i\in [1,N], f\in [1,F]\}$;
          \State $\mathcal{K}=\text{DCT}(\varphi(\mathbf{I}_{LR}))=\{\tau^k_{(t',i,f)}, t'\in [1,t-1], i\in [1,N], f\in [1,F]\}$;
          \State $\mathcal{V}=\text{DCT}(\varphi(\mathbf{I}_{LR}))=\{\tau^v_{(t',i,f)}, t'\in [1,t-1], i\in [1,N], f\in [1,F]\}$;
          \State $R^t = \Lambda_{S}(\tau^q_{(t,i,f)}, \tau^k_{(t,i,f)}, \tau^k_{(t,i,f)})$;
          \State $P^t = \Lambda_{T}(R^t,\Hat{H}^t,\Hat{H}^t)$;
         \State $D^t_{LR} = \text{DCT}(\varphi(I^t_{LR}))$;
         \State $H$ add $\Gamma(D^t_{LR}, P^t)$;
          \State $I_{SR}^t = \text{rDCT}(\Gamma(P^t, D^t_{LR}) + D^t_{LR})$
      \EndFor

  \end{algorithmic}
\end{algorithm}

\begin{figure}[!t]
\centering
\includegraphics[width=\columnwidth]{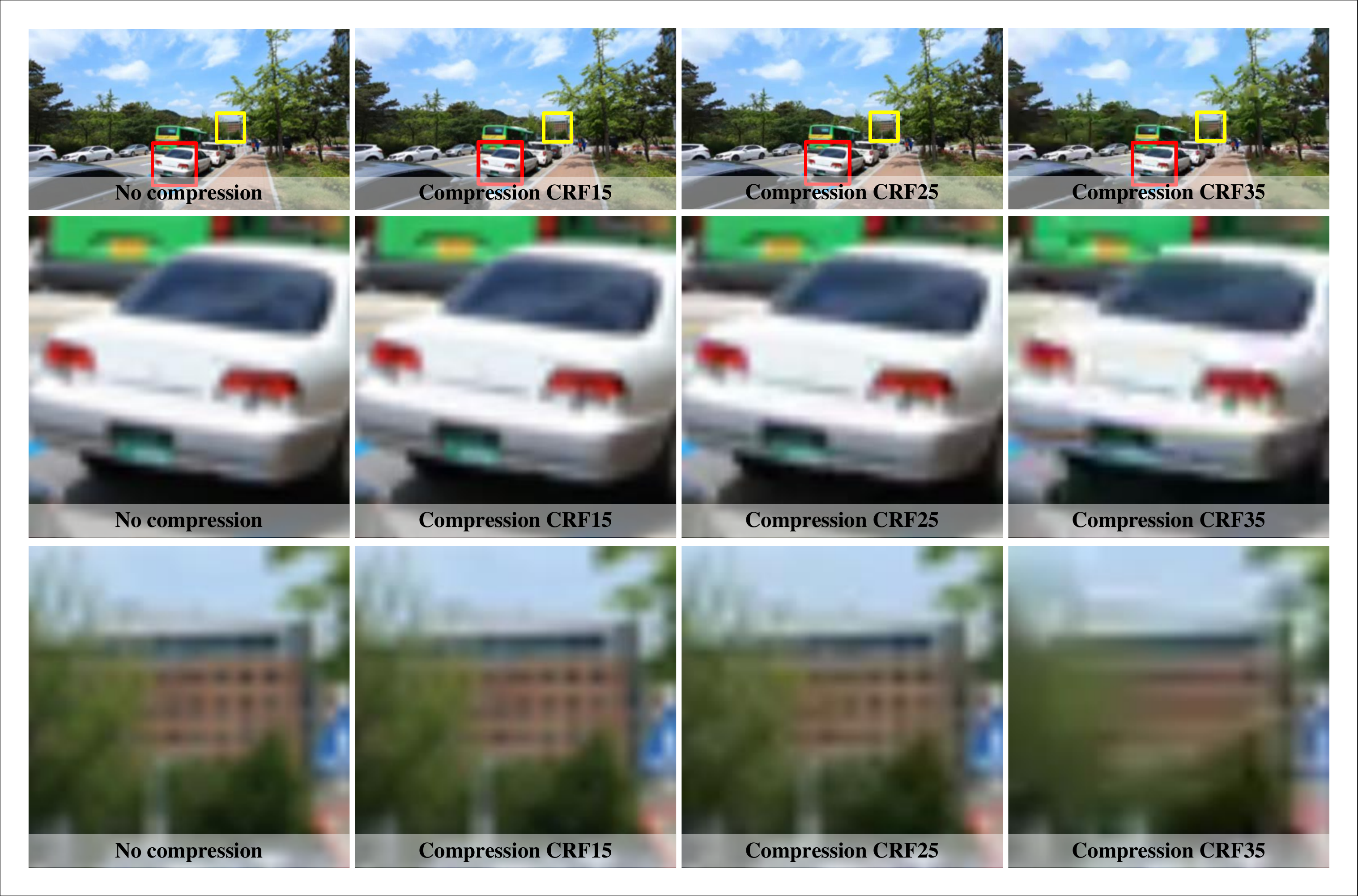}
\caption{The visualization of compressed images with different compression rates}
\label{fig:compress}
\end{figure}

\section{Implemental Details}
\label{sec:implemental}
In this section, we introduce more implemental details that can not be involved in the main paper since the limited pages. All the ablation results are based on the backbone of BasicVSR~\cite{chan2021basicvsr} and final model is based on the backbone of TTVSR~\cite{liu2022learning}. For DCT, we transform image patches of size $B\times B, B=8$ to pixel domain and align them at frequency dimension. Thus, the frequency number is 64. For frequency tokenization, the block size $K\times K$ is $8\times 8$ for a trade-off of computational costs and the performances for different block sizes are shown in Table \ref{tab:bolock}. For COMISR~\cite{li2021comisr}, we retrain it as the settings in \cite{li2021comisr} to generate visualization results.

\begin{table*}[t]
  \caption{The ablation study of block size for space-frequency attention $\Lambda_S$ }
  \centering
  \renewcommand\arraystretch{1.2}
  \renewcommand\tabcolsep{10pt}
  \resizebox{\columnwidth}{!}{
  \begin{tabular}{ c|c|c|c|c}
    \hline

    \hline
     Method & Block Size & CRF15 & CRF25 & CRF35 \\
    \hline
    \multirow{5}{*}{\makecell{FTVSR \\($\Lambda_{S}$)}} & $4\times 4$ &29.62/0.840 & 27.20/0.761 & 24.12/0.646 \\
      &$6\times 6$ & 29.64/0.840 & 27.22/0.761 &24.12/0.645 \\
      &$8\times 8$ &\textbf{29.63/0.840} & \textbf{27.23/0.761} & \textbf{24.12/0.646}\\
      &$12\times 12$ & 29.63/0.840 & 27.22/0.761 & 24.12/0.646 \\
      &$16\times 16$ & 29.61/0.839 & 27.20/0.760 & 24.10/0.644 \\
    \hline
    
    \hline
  \end{tabular}
  }
  \label{tab:bolock}
\end{table*}

\section{Comparison of Compression}
\label{sec:compress}
We follow the same setting as COMISR~\cite{li2021comisr} that adopts ffmpeg to perform the video compression. The compression command is ``ffmpeg -i LR.mp4 -vcodec libx264 -crf CRFvalue save.mp4", where CRF value can be 15, 25, and 35.
The differences of no compression and compression are shown in Figure \ref{fig:compress}. As shown in Figure \ref{fig:compress}, compared with no compression, compression brings more artifacts, which broke the texture structure in the image.

\begin{figure}[!t]
\centering
\includegraphics[width=\columnwidth]{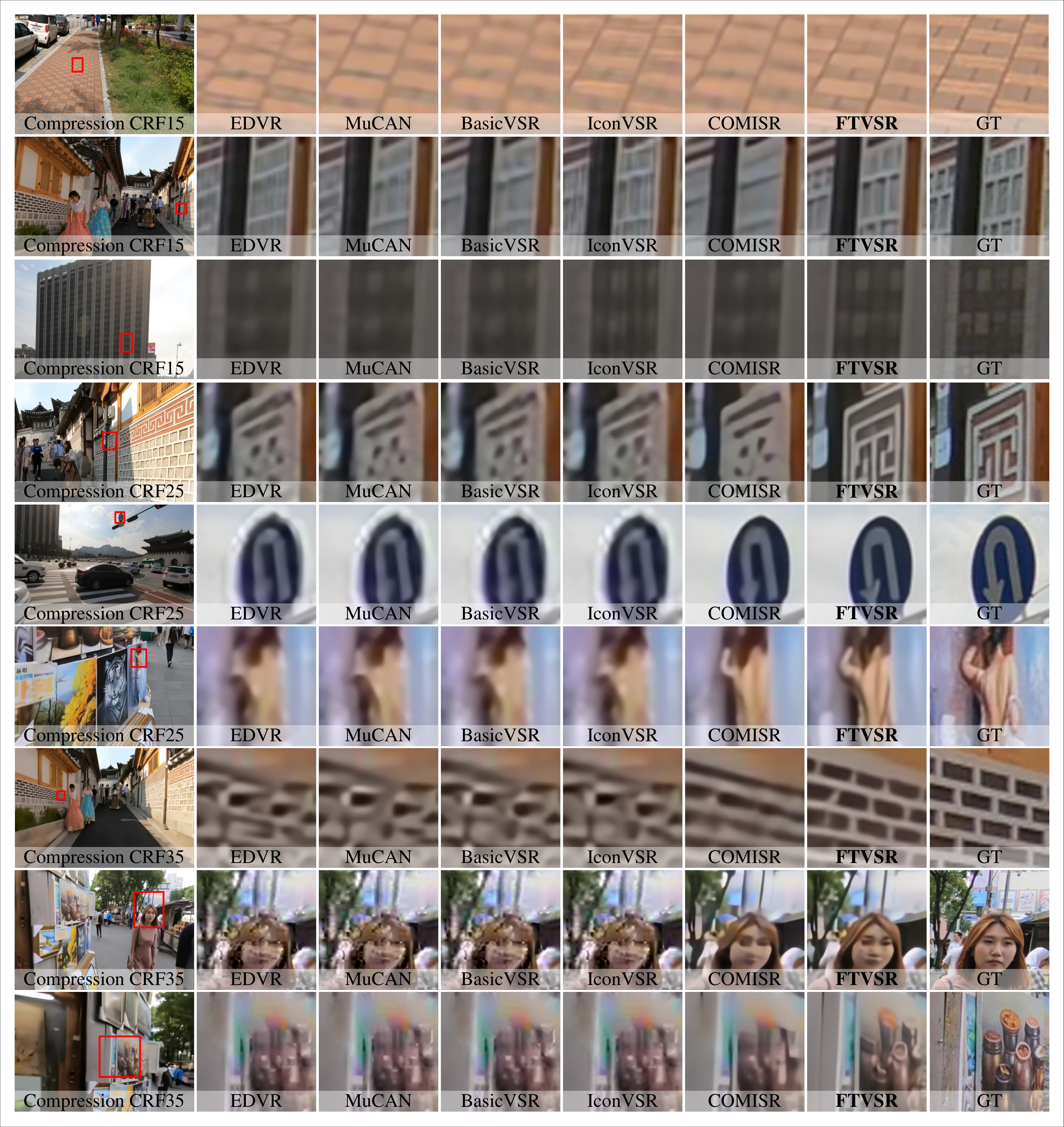}
\caption{The visualization of more results on compressed videos with different compression rates}
\label{fig:morecases}
\end{figure}

\section{Visualization and Failure Cases}
\label{sec:vis}
\subsection{More visualization results}
We compare FTVSR and SOTA methods (EDVR~\cite{wang2019edvr}, MUCAN~\cite{li2020mucan}, BasicVSR~\cite{chan2021basicvsr}, IconVSR~\cite{chan2021basicvsr} and COMISR~\cite{li2021comisr}) on the compressed videos with different compression rates.
The visualization results are shown in Figure \ref{fig:morecases}.

\subsection{Limitations and Failure Cases}
\label{sec:limit}
We discuss the limitations of the proposed FTVSR in this subsection and show some failure cases in Figure \ref{fig:fail}.
\subsubsection{Small Parts}
As shown in the first row of Figure \ref{fig:fail}, some small parts in image, which its textures are relatively small, are not easy to recover since the limited input information.
\subsubsection{Motion Parts}
\begin{figure}[!t]
\centering
\includegraphics[width=\columnwidth]{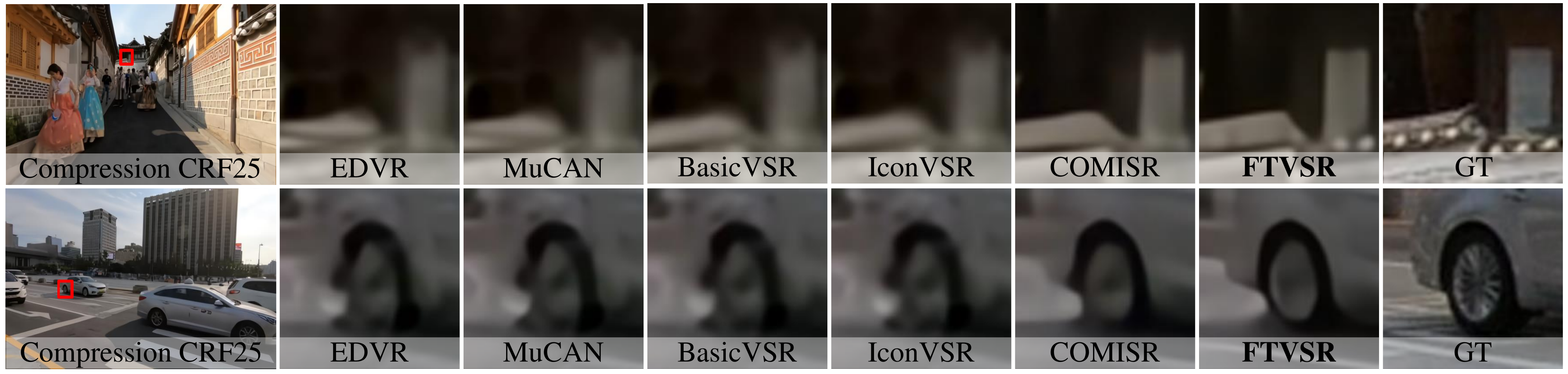}
\caption{The comparison of failure cases with other methods (EDVR~\cite{wang2019edvr}, MUCAN~\cite{li2020mucan}, BasicVSR~\cite{chan2021basicvsr}, IconVSR~\cite{chan2021basicvsr}, COMISR~\cite{li2021comisr}) on compressed videos with compression rate of CRF25}
\label{fig:fail}
\end{figure}

As shown in the second row of Figure \ref{fig:fail}, the textures with complex motion patterns (e.g. rotation) are also not easy to recover. FTVSR can capture the contour texture of the rotating wheel, but fail on the detail texture of the wheel hub.

\end{appendix}

\end{document}